\documentclass{bmvc2k}

\usepackage{times}
\usepackage{epsfig}
\usepackage{graphicx}
\usepackage{algorithm}
\usepackage[noend]{algpseudocode}
\usepackage{amsmath}
\usepackage{amssymb}
\usepackage{verbatim}
\usepackage{color}
\usepackage{multirow}
\usepackage{pifont}

\title{Label Prediction Framework for Semi-Supervised Cross-Modal Retrieval}

\addauthor{Devraj Mandal}{devrajm@iisc.ac.in}{1}
\addauthor{Pramod Rao}{pramodrameshr@iisc.ac.in}{1}
\addauthor{Soma Biswas}{somabiswas@iisc.ac.in}{1}

\addinstitution{
 Department of Electrical Engineering\\
 Indian Institute of Science\\
 Bangalore 560012
}

\runninghead{Devraj}{Label Prediction Framework for Semi-Supervised Learning}

\begin{document}
\maketitle

\begin{abstract}
Cross-modal data matching refers to retrieval of data from one modality, when given a query from another modality.
In general, supervised algorithms achieve better retrieval performance compared to their unsupervised counterpart, as they can learn better representative features by leveraging the available label information. 
However, this comes at the cost of requiring huge amount of labeled examples, which may not always be available.
In this work, we propose a novel framework in a semi-supervised setting, which can predict the labels of the unlabeled data using complementary information from different modalities. 
The proposed framework can be used as an add-on with any baseline cross-modal algorithm to give significant performance improvement, even in case of limited labeled data.
Finally, we analyze the challenging scenario where the unlabeled examples can even come from classes not in the training data and evaluate the performance of our algorithm under such setting.
Extensive evaluation using several baseline algorithms across three different datasets shows the effectiveness of our label prediction framework.
\end{abstract}

\section{Introduction}
For the application of cross-modal retrieval, supervised algorithms \cite{ccca} \cite{seph} \cite{gph} \cite{icip} generally outperform their unsupervised counterparts \cite{cca1} \cite{cca2} \cite{cmfh}, but at the cost of additional label information. 
The performance also greatly depends upon the amount of labeled data \cite{lc1} \cite{lc2} \cite{lc3}. 
Since the task of labeling is often very expensive and time-consuming, designing deep based models to mitigate this shortcoming is also very important~\cite{lc1} \cite{lc2} \cite{lc3} \cite{pseudolabel}.
Semi-supervised learning \cite{ladder} \cite{metembed} \cite{pseudolabel} \cite{deepgen} \cite{semisupembed} treads the middle ground by considering a small subset of data as labeled, and the remaining as unlabeled.
Recently, several semi-supervised cross-modal algorithms have been developed~\cite{adap_unf} \cite{gssl} \cite{semi_hash1} \cite{semi_hash2} \cite{semi_hash3} \cite{semi_dict}, which aims to find out the optimal way to jointly use both the labeled and unlabeled data for getting better performance.
These approaches generally follow one of the three possible strategies namely 
(1) Pre-training the deep network using unlabeled examples followed by training it again with its labeled counterparts, 
(2) Using the unlabeled samples as a regularization term for structure preservation of the embedded features and 
(3) An iterative scheme in which label prediction and network parameter learning are done in an alternate fashion repeatedly.

In this work, given a set of labeled and unlabeled training data in a semi-supervised cross-modal setting, we propose a novel label prediction framework (LPF) to predict the labels for the unlabeled data.
Utilizing the complementary information from both modalities as well as the original features, we filter out the data for which the predicted labels are potentially wrong and select only that portion whose predicted labels are probably correct to re-train the LPF.
These two steps are repeated iteratively and with each iteration, more number of unlabeled examples and their predicted labels are added which helps to train the LPF network better. 
Finally, we use all the labeled and pseudo-labeled examples to train any supervised cross-modal algorithm.
We perform extensive experiments to show the efficacy of our algorithm for different baselines and three datasets, even with limited labeled data. 

In real world, a portion of the unlabeled data can potentially come from novel classes not seen during training.
Here, we analyze the effect of these out-of-class samples on the label prediction and subsequently on the retrieval performance of some baseline cross-modal algorithms.

The main contributions of our work is as follows: \\
{\bf (1)} We propose a novel label prediction framework for predicting labels of unlabeled data in a semi-supervised setting, which can then be fed to any supervised cross-modal algorithm.
{\bf (2)} The proposed framework is effective even in case of limited amount of labeled data. \\
{\bf (3)} We also analyze the effect of out-of-class samples in our approach. To the best of our knowledge this is the first study on the effect of out-of-class samples in a semi-supervised setting for cross-modal retrieval problems. \\
{\bf (4)} Extensive experiments show the usefulness of the proposed framework using several baselines and three different datasets.

Next, we discuss the related work in literature. 
The proposed approach for different scenarios is discussed in Section~\ref{prop}.
The results of experimental evaluation is reported in Section~\ref{expts} and the paper ends with a conclusion.

\section{Related Work}

Here we describe the relevant works in the semi-supervised (SS) setting, first for image classification task and then for cross-modal retrieval. \\ \\
{\bf Semi-supervised image classification:} As discussed in the introduction, three strategies are usually followed in literature for SS scenario.
The first strategy is followed in \cite{deepgen}, but its performance usually suffers since the second stage typically dominates and the model tends to forget what it has learnt in the first stage.
The second strategy is followed in the works of \cite{metembed} \cite{semisupembed} \cite{ent_loss} \cite{semi_gan}. Algorithms like in \cite{ladder} \cite{stacked_ae} employ an hierarchical strategy in which the unlabeled examples are used for image reconstruction and the labeled examples are used for image classification.
The work in \cite{pseudolabel} \cite{lc3} follows the iterative approach of the third strategy. 
\cite{lc3} also has the additional property of growing the network layers if the necessity arises following the accumulation of additional pseudo-labeled examples.
Data augmentation techniques in the image domain can greatly boost image classification performance in the SS setting \cite{temp_ens} \cite{mean_teach}. 
This technique though very useful is difficult to implement for applications in the cross-modal setting. 
\cite{itno} works under the open-set scenario where the labels are noisy and the sample may belong to a class outside of the set of known classes. 
The approach uses the Local Outlier Factor algorithm concurrently with a deep siamese network trained using triplet loss to account for the noisy and out-of-distribution samples. \\ \\
{\bf Semi-supervised cross-modal retrieval:} The problem of SS hashing in the cross-modal setting is relativity less explored and has been addressed in \cite{adap_unf} \cite{gssl} \cite{semi_hash1} \cite{semi_hash2} \cite{semi_hash3} \cite{semi_dict}. In \cite{semi_hash1} \cite{gssl}, multi-graph learning is used over the unlabeled data for structure preservation while learning the common embedding representations. 
The work in \cite{adap_unf} designs a dragging technique with a linear regression model so that embedded features lies close to the correct class labels while pushing the irrelevant samples far apart. 
In \cite{semi_dict}, sparse representation of the different modality data for both the labeled and unlabeled samples are projected into a common domain defined by its class label information. A non-parametric Bayesian approach has been proposed in \cite{semi_hash2} to handle the SS situation.
A novel approach in semi-supervised hashing using Generative Adversarial Network~\cite{semi_hash3} has been used to model the distribution across the different modalities and a loss function has been suitably designed learn to select correct/similar data pairs from the unlabeled set in an adversarial fashion. Though \cite{semi_hash3} shows impressive performance the amount of labeled data required is quite large.

In this work, we propose a novel label prediction framework in a semi-supervised setting which predicts the labels of the unlabeled data, which can then be used to augment the labeled portion of the data and given as input to any baseline cross-modal algorithm.

\section{Proposed Method}
\label{prop}
Here, we describe the proposed framework for both standard semi-supervised setting and the more challenging scenario which also includes out-of-class samples. 
Let the cross-modal data be represented as $\textbf{X}_{t} \in \mathbb{R}^{d_t \times N}$ ($t \in \{1,2\}$), where $t=2$ is the number of modalities, $N$ is the number of training samples and $d_t$ is the feature dimension.
Let the labels be denoted as $\textbf{L} \in \mathbb{R}^{C \times N}$, where $C$ is the number of classes with each sample belonging to a single category.
Consider that the input data $\textbf{X}_t$ consists of (a) $m$ labeled samples denoted by $\textbf{X}_t^l \in \mathbb{R}^{d_t \times m}$ with its corresponding labels $\textbf{L}^l \in \mathbb{R}^{C \times m}$ and (b) $n$ unlabeled samples denoted by $\textbf{X}_t^{ul} \in \mathbb{R}^{d_t \times n}$, with $m+n=N, m \leq n$.
We consider both the labeled and unlabeled data to be paired.
\subsection{Label Prediction Framework}
Given this set of labeled and unlabeled data, first, we describe the Label Prediction Framework (LPF) which is trained to predict the labels of the unlabeled samples.
For training the LPF, we subdivide the labeled portion of the training data $\textbf{X}_t^l$ as $\textbf{X}_t^{tr}, \textbf{L}^{tr}$ and $\textbf{X}_t^{val}, \textbf{L}^{val}$ to form the training and validation sets.
The proposed network architecture is shown in Figure~\ref{figLC} which consists of encoders $\mathcal{E}_t$ and decoders $\mathcal{D}_t$ for both the modalities. 
In our implementation, both $\mathcal{E}_t$ and $\mathcal{D}_t$ consists of three fully connected (fc) layers (in mirror configuration) with ReLU and dropout between all the fc layers, except the final layer. 
The final layer of the encoder has two activation functions, namely (1) softmax for predicting the labels and (2) tanh whose output is subsequently passed through the decoder.
For input data $x_{tj}$ ($j^{th}$ sample from modality $t$) to the encoder $\mathcal{E}_t$, the output of the softmax is denoted as $x_{tj}^{s}$ and that of the tanh layer is denoted as $x_{tj}^{\tanh}$.
The encoded output $x_{tj}^{\tanh}$ is passed through $\mathcal{D}_t$ to get the reconstruction $\hat{x}_{tj}$.
Now, we will describe the different losses used to train this network:

\begin{figure*}[t!]
	\begin{center}		
		\includegraphics[width=0.90\textwidth,height=0.30\textheight]{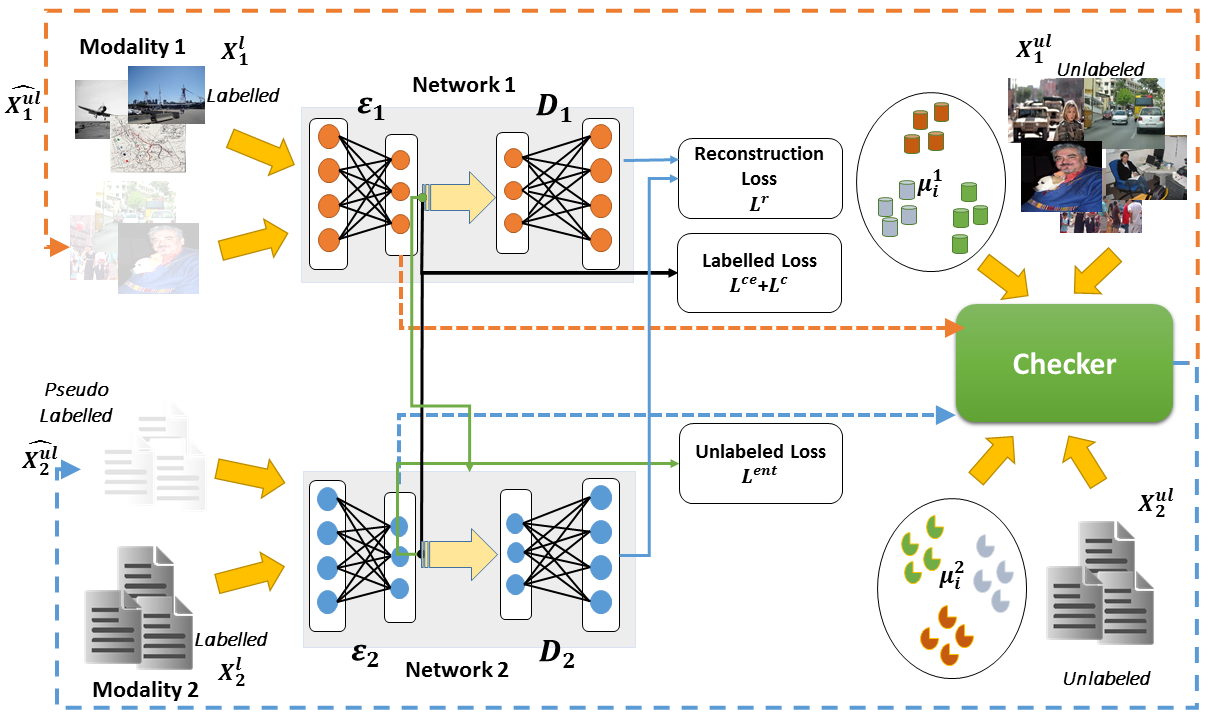}
	\end{center}
	\caption{Illustration of the proposed LPF. Initially, the labeled data $\{\textbf{X}_t^l\}$ is used to learn the parameters of $\{\mathcal{E}_t, \mathcal{D}_t\}$ while minimizing $\mathcal{L}$. 
Next, we use the learned $\mathcal{E}_t$ to predict the labels for the unlabeled data $\textbf{X}_t^{ul}$. The checker constructs the constraint set $\mathcal{C}$ using the original mean features $\mu^t$ and the performance of $\mathcal{E}_t$ on $\textbf{X}_t^{val}$. 
Based on its decision, a subset of the unlabeled data $\hat{\textbf{X}}_t$ is selected and fed back to the network to fine-tune it further (the lightly highlighted elements denotes these pseudo-examples). Thus this model uses complimentary information from the original features as well as the paired information of the cross-modal data to select pseudo-examples judiciously for further fine-tuning of the network.}
	\label{figLC}
\end{figure*}

\begin{enumerate}
	\item {\bf Labeled data:} For the labeled portion of the data, we want the samples from same class to cluster together, which in turn will help in classification. 
We tap the output from the second fc layer in $\mathcal{E}_t$ and denote it as $x_{tj}^f$. We use the following two losses:
\newline  {\bf Cross-entropy loss:} $\mathcal{L}^{ce}=- \sum_{j=1}^{m} \log \left( e^{x_{tj}^{s}[i]} / \sum_{k=1}^{C} e^{x_{tj}^{s}[k]} \right)$ is used to minimize the classification errors over the labeled examples. Here, $i$ is the correct class index.
\newline  {\bf Center loss} \cite{cent_loss}: We use this loss to minimize the distance of each sample with respect to its center representation as follows: $\mathcal{L}^{c} = \sum_{j=1}^{m} \sum_{k=1}^{C} \mathbf{1}_{\{L_{j}=k\}} \vert \vert x_{tj}^f - c_{t,k}\vert \vert_2^2$.
This also ensures that the samples from the same class are clustered together.
$c_{t,k}$ denotes the $k^{th}$ class center for $t^{th}$ modality and $\mathbf{1}_{\{L_{j}=k\}}$ is the indicator variable which gets activated when label $L_{j}$ of the sample $x_{tj}$ is consistent with the correct center. 
The centers $\{c_{t,k}\}_{k=1}^C$ are learned while training the network using~\cite{cent_loss}.
We consider learning the centers $c_{t,k}$ from the second fc layer output as the final layer length is limited by the number of training categories, which is often small and hence the learned center representations might not be discriminative enough. 
The two losses are important~\cite{cent_loss}, since $\mathcal{L}^{c}$ helps to make the classification using $\mathcal{L}^{ce}$ better by pushing the centers apart and making the features of each individual classes as clustered together as possible.

	\item {\bf Unlabeled data:} To make the label predictions of the unlabeled data less ambiguous, we utilize the {\bf Entropy Regularization loss} \cite{ent_loss} as $\mathcal{L}^{ent} = \sum_{j=m+1}^{N} - x_{tj}^s \log (x_{tj}^s)$.
Since the unlabeled samples belong to one of the $C$ categories, we want to make the softmax probability for a particular class as high as possible, which in turn is equivalent to minimizing the entropy of the prediction. 

	\item {\bf Labeled and Unlabeled data:} To ensure that there is no loss of information in the encoder-decoder structure, for both the labeled and unlabeled samples, we use a {\bf Reconstruction loss} at the decoder output given as $\mathcal{L}^{r} = \sum_{j=m+1}^{N} \vert \vert x_{tj}-\hat{x}_{tj} \vert \vert_2^2 $.
\end{enumerate} 
Thus, the total loss function for the entire network is given as: 
$\mathcal{L} = \alpha^{ce} \mathcal{L}^{ce} +  \alpha^{c} \mathcal{L}^{c} + + \alpha^{ent} \mathcal{L}^{ent} + \alpha^{r} \mathcal{L}^{r}$ where $\alpha^{ce}, \alpha^{c}, \alpha^{ent}, \alpha^{r}$ are the tunable hyper-parameters.
Once the LPF network is trained, we can use it to predict the labels for the unlabeled samples. \\ 
{\bf Exploiting paired information in cross-modal data:} In this work, we leverage the complementary information available in the paired unlabeled data of the two modalities to verify if the label prediction given by LPF is reliable.
For the $j^{th}$ unlabeled data $x_{tj}$, if $x_{tj}^s$ is the softmax output from $\mathcal{E}_t$, the predicted label is given by $l_{tj}^{\mathcal{E}_t} = \arg \max_i  x_{tj}^s[i]; \hspace{5 pt} 0 \leq i \leq (C-1)$.
At the beginning of training, due to very limited amounts of data, these predictions are not reliable. 
This can be partially corrected by studying how close the original features are to their mean feature representations. 
Utilizing this fact, an alternate prediction on the unlabeled data can be made and both the predictions can be combined suitably to select the reliable predictions.
Let us denote the mean features of each class for the $t^{th}$ modality as $\{\mu_t^1,...,\mu_t^{C}\}$. 
The means are computed using the original feature representation of the labeled data $X_t^{l}$.
The closest distance of the sample $x_{tj}$ to this mean feature set is also a coarse prediction of the class it belongs to and is given by
$ l_{tj}^{\mu} = \arg \min_i \vert \vert x_{tj}-\mu_t^i \vert \vert_2^2$.
These four predictions can be computed for each data pair in $\mathbf{X}_t^{ul}$.
Finally, the correctness of the label prediction is verified using a threshold $\tau$.
Specifically, a data pair $(x_{1j},x_{2j})$ can be assumed to be correctly predicted if it satisfies the following conditions
\begin{equation}
\textbf{(1)} x_{1j}^s[i] \geq \tau, \qquad \textbf{(2)} x_{2j}^s[i] \geq \tau, \qquad \textbf{(3)} l_{1j}^{\mu} = l_{1j}^{\mathcal{E}_1}, \qquad \textbf{(4)} l_{2j}^{\mu} = l_{2j}^{\mathcal{E}_2}
\end{equation}
Let us term these set of constraints as $\mathcal{C}$.
Here, we set $\tau = 0.9$ for all our experiments.
This essentially implies that the confidence of the network's predictions must be more than $\tau$ each and it must match with the predictions made by the original features individually.

Since the two modalities have different features which may have different discriminative ability, we use a more relaxed condition to determine the correctness of the label prediction by taking either condition (\textbf{(1)} \& \textbf{(3)}) or (\textbf{(2)} \& \textbf{(4)}).
The choice between the two conditions depends on the performance of $\mathcal{E}_t$ on the validation set $\textbf{X}_t^{val}$.
Let the accuracy of $\mathcal{E}_1,\mathcal{E}_2$ on $\textbf{X}_1^{val}, \textbf{X}_2^{val}$ by denoted as $cf_1$ and $cf_2$. Then the set $\mathcal{C}$ is determined as follows
\begin{eqnarray}
\small
\text{if, $cf_1$} \geq \text{$cf_2$, \hspace{5 pt} $\Rightarrow$ $\mathcal{C}=\{\textbf{(1)},\textbf{(3)}\}$} \hspace{15 pt} \text{otherwise, \hspace{2 pt} $\mathcal{C}=\{\textbf{(2)},\textbf{(4)}\}$} \nonumber
\end{eqnarray}
This approach has two fold advantages, (1) It automatically selects the good features, thus there is no need for manual intervention and (2) automatic switching may occur which basically means that the label predictions will be driven in a complimentary fashion.
In addition, since we are updating the constraint set $\mathcal{C}$ at each iteration to reflect the better classifier's performance, it is expected that increasingly more number of correctly labeled examples from $\textbf{X}_t^{ul}$ gets selected.
We thus get the set of unlabeled examples whose predictions are likely to be correct as
$\hat{\textbf{X}}_t^{ul} = \{ x \hspace{5 pt} | \hspace{5 pt} x \in \textbf{X}_t^{ul} \hspace{5 pt} \text{and satisfies} \hspace{5 pt} \mathcal{C} \}$.

Now, the expanded labeled set is given by $\hat{\textbf{X}}_t = [\textbf{X}_t^{tr} \hspace{5 pt} \hat{\textbf{X}}_t^{ul}]$ with labels $\hat{\textbf{L}} = [\textbf{L}^{tr} \hspace{5 pt} \textbf{L}^{pl}]$, where $\textbf{L}^{pl}$ are the predicted labels.
We use this data to further fine-tune our LPF network with a smaller learning rate. 
We repeat the label prediction and network fine-tuning iteratively until the cardinality of $\hat{\textbf{X}}_t^{ul}$ saturates.
Finally, we can feed $( \hat{\textbf{X}}_t, \hat{\textbf{L}} )$ to any supervised cross-modal baseline algorithm for retrieval.
Algorithm 1 gives the different steps of the LPF.

\begin{algorithm}[!t]
	\label{Algo_LC}
	\caption{The Label Prediction Network} \label{algo1}
	\begin{algorithmic}[1]
		\State \textbf{Input} : $\textbf{X}_t^{tr}, \textbf{X}_t^{val}, \textbf{X}_t^{ul} \hspace{2 pt} \{t=1,2\},  \textbf{L}^{tr}, \textbf{L}^{val}$.
		\State \textbf{Output} : Data $\hat{\textbf{X}}_t$ and their predicted labels $\hat{\textbf{L}}$.
		\State \textbf{Initialize} : Initialize the network parameters of $\mathcal{E}_t, \mathcal{D}_t$. Learn the mean feature sets $\mu^t$.
		\State Train the classifiers $\mathcal{E}_t, \mathcal{D}_t$ using ($\textbf{X}_t^{tr}, \textbf{L}^{tr}$) by computing the loss $\mathcal{L}$ and back-propagating the error.
		\State Continue until $|\hat{\textbf{X}}_t^{ul}|$ does not change or until $T$ iterations (whichever earlier):
		\State \hspace{5 pt} Measure performance $cf_t$ on validation set $\textbf{X}_t^{val}, \textbf{L}^{val}$ using $\mathcal{E}_t$.
		\State \hspace{5 pt} Determine $l_{tj}^{\mu}, l_{tj}^{\mathcal{E}^t}$ for each sample in unlabeled set $\textbf{X}_t^{ul}$.
		\State \hspace{5 pt} Construct the new constraint set $\mathcal{C}$. Use this to determine $\hat{\textbf{X}}_t^{ul}$.
		\State \hspace{5 pt} Form $\hat{\textbf{X}}_t$ as $[\textbf{X}_t^{tr} \hspace{5 pt} \hat{\textbf{X}}_t^{ul}]$ \& $\hat{\textbf{L}}$ as $[\textbf{L}^{tr} \hspace{5 pt} \textbf{L}^{pl}]$.
		\State \hspace{5 pt} Fine-tune $\mathcal{E}_t, \mathcal{D}_t$ with a lower learning rate to update the network parameters.
	\end{algorithmic}
\end{algorithm}

\section{Experimental Evaluation}
\label{expts}
We consider three standard single label datasets for evaluating the proposed approach.
The UCI Digit data~\cite{uci} contains different feature representations of handwritten numerals for ten categories i.e., ($0$-$9$) with $200$ examples each. The train:test split is $1500$:$500$. The features used for our experiments are the same as in \cite{fsh}.
The LabelMe data~\cite{labelme} contains image-text pairs from eight different categories. GIST features are considered for the image domain and Word frequency vector for the text domain \cite{labelme}. We take $200$ samples from each category in the training set and the rest of the samples in the testing set.
Wiki data~\cite{wiki} contains $2,866$ image-text pairs from $10$ different categories.
The images and texts are represented using $4096$-d CNN descriptors and $100$-d word vectors respectively.
The train:test split considered is $2000$:$866$ as in~\cite{gssl}.

Mean Average Precision (MAP) is used as our evaluation metric for baseline comparisons against the other the cross-modal retrieval methods. It is defined as the mean of the average precision (AP) for all queries. Average Precision can be defined as $AP(q) = \frac{\sum_{r=1}^{R} P_q(r) \delta(r)}{\sum_{r=1}^{R} \delta(r)}$, where $q$ is the query element and $R$ is the retrieval set. The precision for query $q$ at position $r$ is denoted as $P_q(r)$. MAP@R essentially measures the retrieval accuracy when $R$ number of items from the database are being retrieved per query item. We report MAP@50 for all our experiments \cite{gssl}.

We consider a variety of baseline cross-modal algorithms like CCCA \cite{ccca} GSSL \cite{gssl} GsPH \cite{gph} LCMF \cite{icip} SCM \cite{scmh} (SCM$_s$ and SCM$_o$ denotes the sequential and orthogonal versions) SePH \cite{seph} SMFH \cite{smfh} ACMR \cite{acmr} GrowBit \cite{growbit} with which we integrate our LPF. 
We take the publicly available versions of the author's codes or re-implement them wherever necessary while running the baseline algorithms. We set the parameters of the baseline algorithms in accordance to strategies described in the individual papers. 
For the hashing based approaches, we used hash code of length $64$.

\subsection{Results for Semi-Supervised Protocol}
Here, we report the results of our LPF module as an add-on with other baseline approaches for the three datasets. 
We denote the results for each algorithm under three different modes, 
(1) `f'', denotes that the algorithm is working in supervised mode with no unlabeled data ;
(2) ``l'', using only labeled portion of the data and 
(3) ``ss'', where the pseudo-labeled examples as predicted by LPF are provided in addition to the labeled data. 
We consider $\rho \%$ of the total training data as labeled, and the remaining as unlabeled and we report results for $\rho = \{10\%, 30\%, 50\%\}$.
All experiments are repeated over $5$ random labeled:unlabeled split and the average results are reported in Table \ref{semi_results1}.
We make the following observations, 
(1) the result of ``f'' mode is the best as expected as it has access to all the labeled training data; 
(2) the results under ``ss'' mode is better than ``l'' mode which signifies that the proposed LPF is able to correctly predict the labels of the unlabeled set and pass it to the baseline algorithms; 
(3) the importance of LPF module is more when $\rho$ is low, i.e. when the amount of labeled data is very limited, thus making the training more challenging;
(4) LPF works equally well for non-deep and deep based algorithms. 
We observe similar pattern as we increase $\rho$ from $50\%$ to $90\%$, though the performance difference between the ``l'' and ``ss'' is less.

\begin{table}[t!]
	\footnotesize	
	\renewcommand{\arraystretch}{1.0}
	\setlength{\tabcolsep}{5.5 pt}
	\centering
	\caption{Average MAP@50 on UCI \cite{uci}, Wiki \cite{wiki} and LabelMe \cite{labelme} datasets. Here, ``f'', ``l'' and ``ss'' denotes the three modes of operation. $^+$ indicates deep based algorithms. $^*$ indicates that GSSL is working in a semi-supervised mode in ``b'' and ``c''.}
	\begin{tabular}{|c|c|c c c|c c c|c c c|}
		\hline
		&  & \multicolumn{3}{c|}{UCI \cite{uci}} & \multicolumn{3}{c|}{Wiki \cite{wiki}} & \multicolumn{3}{c|}{LabelMe \cite{labelme}} \\ \hline
		& $\rho$ & 10 & 30 & 50 & 10 & 30 & 50 & 10 & 30 & 50 \\ \hline
		\multirow{3}{*}{CCCA \cite{ccca}} & f & \multicolumn{3}{c|}{0.667} & \multicolumn{3}{c|}{0.419} & \multicolumn{3}{c|}{0.639} \\ \cline{3-11}
		& l & 0.634 & 0.648 & 0.657 & 0.314 & 0.362 & 0.381 & 0.573 & 0.620 & 0.628 \\ 
		& ss & 0.639 & 0.655 & 0.657 & 0.379 & 0.398 & 0.400 & 0.611 & 0.624 & 0.640 \\ \hline		
		\multirow{3}{*}{GsPH \cite{gph}} & f & \multicolumn{3}{c|}{0.853} & \multicolumn{3}{c|}{0.473} & \multicolumn{3}{c|}{0.820} \\ \cline{3-11} 
		& l & 0.779 & 0.821 & 0.833 & 0.359 & 0.426 & 0.451 & 0.717 & 0.784 & 0.794 \\
		& ss & 0.800 & 0.833 & 0.842 & 0.426 & 0.449 & 0.467 & 0.746 & 0.792 & 0.792 \\ \hline
		\multirow{3}{*}{LCMF \cite{icip}} & f & \multicolumn{3}{c|}{0.847} & \multicolumn{3}{c|}{0.484} & \multicolumn{3}{c|}{0.827} \\ \cline{3-11} 
		& l & 0.774 & 0.819 & 0.834 & 0.354 & 0.422 & 0.447 & 0.719 & 0.790 & 0.799 \\
		& ss & 0.809 & 0.830 & 0.843 & 0.425 & 0.451 & 0.470 & 0.758 & 0.799 & 0.801 \\ \hline
		\multirow{3}{*}{SCM$_s$ \cite{scmh}} & f & \multicolumn{3}{c|}{0.652} & \multicolumn{3}{c|}{0.358} & \multicolumn{3}{c|}{0.694} \\ \cline{3-11} 
		& l & 0.509 & 0.584 & 0.595 & 0.274 & 0.313 & 0.328 & 0.554 & 0.630 & 0.681 \\
		& ss & 0.598 & 0.628 & 0.626 & 0.332 & 0.328 & 0.331 & 0.634 & 0.651 & 0.676 \\ \hline
		\multirow{3}{*}{SCM$_o$ \cite{scmh}} & f & \multicolumn{3}{c|}{0.437} & \multicolumn{3}{c|}{0.273} & \multicolumn{3}{c|}{0.475} \\ \cline{3-11} 
		& l & 0.364 & 0.403 & 0.402 & 0.214 & 0.236 & 0.239 & 0.382 & 0.423 & 0.453 \\
		& ss & 0.401 & 0.411 & 0.430 & 0.234 & 0.243 & 0.240 & 0.420 & 0.432 & 0.437 \\ \hline
		\multirow{3}{*}{SePH \cite{seph}} & f & \multicolumn{3}{c|}{0.844} & \multicolumn{3}{c|}{0.477} & \multicolumn{3}{c|}{0.813} \\ \cline{3-11}
		& l & 0.781 & 0.820 & 0.834 & 0.359 & 0.429 & 0.454 & 0.717 & 0.774 & 0.787 \\
		& ss & 0.800 & 0.827 & 0.833 & 0.429 & 0.453 & 0.461 & 0.740 & 0.790 & 0.790 \\ \hline
		\multirow{3}{*}{SMFH \cite{smfh}} & f & \multicolumn{3}{c|}{0.686} & \multicolumn{3}{c|}{0.335} & \multicolumn{3}{c|}{0.711} \\ \cline{3-11} 
		& l & 0.584 & 0.651 & 0.659 & 0.277 & 0.301 & 0.315 & 0.580 & 0.649 & 0.688 \\
		& ss & 0.654 & 0.667 & 0.662 & 0.321 & 0.322 & 0.324 & 0.684 & 0.703 & 0.709 \\ \hline
		\multirow{3}{*}{ACMR$^+$ \cite{acmr}} & f & \multicolumn{3}{c|}{0.776} & \multicolumn{3}{c|}{0.444} & \multicolumn{3}{c|}{0.828} \\ \cline{3-11}
		& l & 0.543 & 0.694 & 0.721 & 0.319 & 0.396 & 0.411 & 0.642 &  0.765 & 0.801 \\
		& ss & 0.751 & 0.757 & 0.768 & 0.421 & 0.440 & 0.436 & 0.767 & 0.797 & 0.823 \\ \hline
		\multirow{3}{*}{GrowBit$^+$ \cite{growbit}} & f & \multicolumn{3}{c|}{0.812} & \multicolumn{3}{c|}{0.465} & \multicolumn{3}{c|}{0.833} \\ \cline{3-11}
		& l & 0.558 & 0.773 & 0.784 & 0.279 & 0.390 & 0.419 & 0.654 & 0.792 & 0.812 \\
		& ss & 0.785 & 0.795 & 0.802 & 0.409 & 0.445 & 0.448 & 0.756 & 0.796 & 0.816 \\ \hline \hline
		\multirow{3}{*}{GSSL$^*$ \cite{gssl}} & f & \multicolumn{3}{c|}{0.731} & \multicolumn{3}{c|}{0.455} & \multicolumn{3}{c|}{0.739} \\ \cline{3-11}
		& b & 0.429 & 0.535 & 0.566 & 0.180 & 0.226 & 0.229 & 0.373 & 0.385 & 0.430 \\
		& c & 0.589 & 0.588 & 0.589 & 0.254 & 0.257 & 0.247 & 0.411 & 0.398 & 0.424 \\ \hline
	\end{tabular}
\label{semi_results1}
\end{table}

We conduct an additional experiment with the state-of-the-art semi-supervised approach GSSL \cite{gssl}. 
In Table \ref{semi_results1}, for GSSL, ``b'' implies that all labeled and unlabeled samples are provided to the algorithm, and ``c'' implies the case where it uses the labeled data, LPF predicted pseudo-labeled data and the remaining unlabeled data. 
In ``b'' and ``c'', GSSL is working as a semi-supervised algorithm. 
Though GSSL is designed to handle unlabeled data, in this case also, the proposed LPF gives significant improvement justifying its usefulness.

\begin{figure*}[h!]
	\begin{center}		
		\includegraphics[width=1.0\textwidth,height=0.20\textheight]{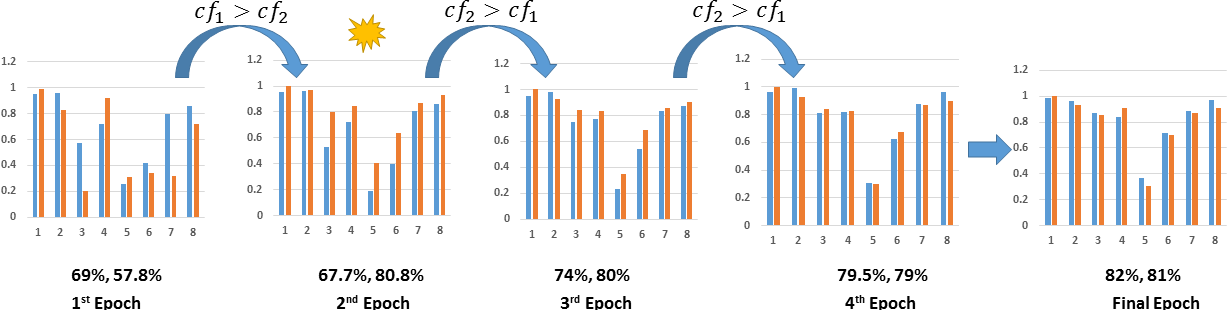}
	\end{center}
	\caption{Here, we show how the two networks are inter-playing among each other to help learn better. In each plot, from left to right, the red and blue bars ($8$ to denote each category) indicate per-class accuracy over the unlabeled data for the a single split of LabelMe \cite{labelme} dataset with $10\%$ provided labels. We observe that both the networks are learning better as the initial accuracy of $\{69\%,57.8\%\}$ has increased to $\{82\%,81\%\}$ at the end of the run. The star indicates the automatic switching phenomenon.}
	\label{figLE}
\end{figure*}

Now we show how the class prediction accuracies for the unlabeled data is evolving with each epoch for the LabelMe dataset \cite{labelme} with $10\%$ labeled data in Figure \ref{figLE}. The red and blue bars denote the per class accuracy over the unlabeled data of the two modalities (the higher the better). We also denote the average accuracy over all the $8$ categories in the LabelMe \cite{labelme} dataset below each bar chart in Figure \ref{figLE}. We observe that from left to right, the accuracy of both the networks improve. The phenomenon of automatic switching where the constraint set is changing to reflect the better updated network is shown with a star. Automatic switching helps the network to interplay among themselves for a better learning mechanism.
 
Table \ref{semi_per_epoch} reports how many examples are being selected per epoch as the algorithm proceeds, and the accuracy of selection of the unlabeled examples for LabelMe \cite{labelme} and UCI \cite{uci} data. 
We observe that as the algorithm proceeds, more number of examples gets selected and with good accuracy. The extra correct examples in addition helps to give better results when baseline algorithms are run with these predicted labels.

\begin{table}[h!]
	\small	
	\renewcommand{\arraystretch}{1.0}
	\setlength{\tabcolsep}{3.0 pt}
	\centering
	\caption{This table reports the number of examples that are selected per epoch to be fed back to $\{\mathcal{E}_t, \mathcal{D}_t\}$ for further fine-tuning. We also report the prediction accuracy of the selected examples per epoch (the higher the better). This is observed on the UCI \cite{uci} and LabelMe \cite{labelme} datasets with $10\%$ labeled data.}
	\begin{tabular}{|c|c|c|c|c|c|}
		\hline
		& 1st Epoch & 3rd Epoch & 5th Epoch & 7th Epoch & 9th Epoch \\ \hline
		UCI \cite{uci} & (730, 96.98\%) & (772,96.63\%) & (817, 96.2\%) & (887, 96.50\%) & (910, 96.59\%) \\ \hline
		LabelMe \cite{labelme} & (971, 95.7\%) & (997, 95.58\%) & (1025, 95.70\%) & (1027, 95.7\%) & (1040, 95.28\%) \\ \hline
	\end{tabular}
\label{semi_per_epoch}
\end{table}

\subsection{Results in Presence of Novel Class Samples}

Here we analyze the performance of the baseline algorithms with the LPF module when the unlabeled data consist of out-of-class samples~\cite{itno}. 
This scenario can occur as stated in \cite{itno} where provided examples from the web might not belong to any of the training class set. 
To simulate this scenario, we create $5$ random splits, in which we divide the $C$ categories into $C^s$ and $C^{us}$ sets. 
We divide the $\{10, 8\}$ categories of the Wiki \cite{wiki} and LabelMe \cite{labelme} into $\{C^s:C^{us}\}$ set of $\{7:3,5:3\}$ classes respectively. 
Here, the training data consist of the labeled examples $\textbf{X}_t^l$ with their labels $\textbf{L}^l \in \mathbb{R}^{C^s \times m}$, and the unlabeled data $\textbf{X}_t^{ul}$ whose labels belongs to $C^s \cup C^{us}$.
Let the amount of in-class:out-of-class data in $\textbf{X}_t^{ul}$ occur in ratio of $1:\kappa$. 
We set $\kappa=\{0.5, 1.5\}$ and re-evaluate the four baseline algorithms and analyze their performance in Table~\ref{semi_out_results} for the two datasets.
We report the results using ``f'': full supervised data, ``l'': only labeled data and ``ss'': semi-supervised mode.

\begin{table}[h!]
	\footnotesize	
	\renewcommand{\arraystretch}{1.0}
	\setlength{\tabcolsep}{7.5 pt}
	\centering
	\caption{Average MAP@50 results on Wiki \cite{wiki} and LabelMe \cite{labelme} datasets when out-of-class samples are present in $\textbf{X}_t^{ul}$. ``f'', ``l'' and ``ss'' denote the different modes of operation.}
	\begin{tabular}{|c|c|c c|c c|c c|c c|}
		\hline
		&  & \multicolumn{4}{c|}{Wiki} & \multicolumn{4}{c|}{LabelMe} \\ \hline
		&  & \multicolumn{2}{c|}{$\kappa=0.5$} & \multicolumn{2}{c|}{$\kappa=1.5$} & \multicolumn{2}{c|}{$\kappa=0.5$} & \multicolumn{2}{c|}{$\kappa=1.5$} \\ \hline
		& $\rho$ & 10 & 30 & 10 & 30 & 10 & 30 & 10 & 30 \\ \hline
		\multirow{3}{*}{CCCA \cite{ccca}} & f & \multicolumn{4}{c|}{0.418} & \multicolumn{4}{c|}{0.767} \\ \cline{3-10}
		& l & 0.351 & 0.400 & 0.353 & 0.384 & 0.741 & 0.746 & 0.681
		 & 0.749 \\
		& ss & 0.367 & 0.392 & 0.385 & 0.393 & 0.667 & 0.741 & 0.617
		& 0.699 \\ \cline{1-10} 
		\multirow{3}{*}{GsPH \cite{gph} } & f & \multicolumn{4}{c|}{0.494} & \multicolumn{4}{c|}{0.881} \\ \cline{3-10}
		& l & 0.397 & 0.461 & 0.415 & 0.462 & 0.815 & 0.856 & 0.844
		& 0.890 \\
		& ss & 0.410 & 0.474 & 0.424 & 0.474 & 0.808 & 0.864 & 0.820
		& 0.874 \\ \cline{1-10}
		\multirow{3}{*}{LCMF \cite{icip} } & f & \multicolumn{4}{c|}{0.499} & \multicolumn{4}{c|}{0.891} \\ \cline{3-10}
		& l & 0.401 & 0.459 & 0.412 & 0.462 & 0.821 & 0.870 & 0.854
		& 0.884 \\
		& ss & 0.409 & 0.469 & 0.426 & 0.479 & 0.812 & 0.874 & 0.821
		& 0.878 \\ \cline{1-10}
		\multirow{3}{*}{SMFH \cite{smfh}} & f & \multicolumn{4}{c|}{0.368} & \multicolumn{4}{c|}{0.867} \\ \cline{3-10}
		& l & 0.315 & 0.348 & 0.323 & 0.346 & 0.781 & 0.828 & 0.751
		& 0.842 \\
		& ss & 0.334 & 0.358 & 0.349 & 0.349 & 0.789 & 0.859 & 0.783
		& 0.858 \\ \cline{2-10} \hline
	\end{tabular}
\label{semi_out_results}
\end{table}

We investigate the results and draw the following conclusions: (1) Though our method LPF in Table \ref{semi_results1} gave significant improvements over the ``l'' mode of operation, here we observe that the performance suffers due to the inclusion of the out-of-class samples;
(2) Interestingly, we observe that for the Wiki \cite{wiki} dataset, the proposed LPF model gives better performance as compared to the ``l'' mode in most cases. 
This is probably because CNN features are being used for the Wiki \cite{wiki} dataset, as compared to the handcrafted features for the LabelMe data \cite{labelme}.
(3) SMFH \cite{smfh} was found to work well under this scenario for both datasets.
(4) A potential way to mitigate this performance degradation is the inclusion of a novel-class sample detector along with the LPF module. \\ \\
\textbf{Implementation Details:} The $\mathcal{E}_t$ in LPF module has 3 fc layers of size $250-250-C$ with $\mathcal{D}_t$ having the mirror architecture. 
We train $(\mathcal{E}_t, \mathcal{D}_t)$ using Stochastic Gradient Descent with learning rate between $lr=10^{-2}-10^{-3}$ for $200$ epochs respectively with early stopping condition.
LPF fine-tuning is done using a lower learning rate typically $lr=10^{-4}-10^{-5}$. For updating the centers, a learning rate of $lr'=5 lr$ is used. The hyper parameters for the loss function are set as $\alpha^{ce}=1, \alpha^r=0.01, \alpha^{c}=0.5$ and $\alpha^{ent}=1$. 

\section{Conclusion and Future Work}
In this work, we propose a novel label prediction framework in semi-supervised setting which can act as an add-on to any cross-modal retrieval baseline algorithm to achieve better performance even in case of limited labeled data. 
Experimental observations have shown that the proposed LPF works equally for both deep and non-deep based methods.
We also analyzed the proposed algorithm for scenarios, where the unlabeled data can potentially contain out-of-class samples.
To the best of our knowledge, this is the first study on the effects of novel class samples on cross-modal retrieval performance, and our analysis indicate that integrating a suitable novel class detector with the LPF can be a future research direction.

\bibliography{egbib}
\end{document}